\documentclass{IEEEtran}
\usepackage{graphicx}
\usepackage{balance}
\usepackage{xcolor}
\usepackage{cite}
\usepackage{array}
\usepackage{mathtools}

\begin{document}

\title{A Single-Layer Asymmetric RNN: Potential Low Hardware Complexity Linear Equation Solver}

\author{Mohammad Samar Ansari \\ 
\thanks{Author is with the Aligarh Muslim University, India.  (Email:  samar.ansari@zhect.ac.in.) 
}}

\IEEEpubid{\begin{minipage}{\textwidth}\ \\ \\ \\ \\ [12pt]
\textit{Preprint submitted to Neurocomputing}.
\end{minipage}}

\markboth{}%
{}

\maketitle

\begin{abstract}
    A single layer neural network for the solution of linear equations is presented. The proposed circuit is based on the standard Hopfield model albeit with the added flexibility that the interconnection weight matrix need not be symmetric. This results in an asymmetric Hopfield neural network capable of solving linear equations. PSPICE simulation results are given which verify the theoretical predictions. Experimental results for circuits set up to solve small problems further confirm the operation of the proposed circuit. 
\end{abstract}


\section{Introduction}
\label{sle-hnn:Introduction}

Efficient and fast solution of a system of linear equations (LE) has been in focus of research for quite some time now. With the advent of digital computers, algorithms took center-stage for such tasks. However, even modern day computers running highly optimized algorithms do not match up to the speeds offered by analog computation for providing the solution for systems of simultaneous linear equations (SLE) \cite{xiao2019nonlinear,jin2018noise,ansari2011dvcc,rahman2011neural}.

The Hopfield neural network (also popular by the more simplified `Hopfield Network') is widely known to be amongst the most prominent and influential models in the research on bio-inspired neural architectures and artificial neural networks (ANNs) \cite{Xu96}. Several other competing architectures that have appeared in the technical literature following the rise and popularity of Hopfield networks, for instance Boltzmann machine (also called stochastic Hopfield network) \cite{Ack85}, bi-directional associative memory (BAM) \cite{Kos88},  attractor neural network with Q-state neurons \cite{Koh93}, etc., have their origins in the theory of the Hopfield Network, and therefore are either improved generalizations or direct minor/major variants of the standard Hopfield network \cite{Xu96}. In fact, a quick survey of the technical literature leads to the observation that thehe Hopfield neural network (HNN) has indeed been found to be quite useful in an assortment  of applications, such as face recognition, solution of the Travelling Salesman Problem (which finds wide applicability in a range of different scenarios), Content Addressable Memories (CAM), analog-to-digital (A/D) conversion, solution of linear programming problems (LPP), grayscale image recognition, and signal decision. To embed and solve problems related to signal processing, grayscale image recognition, neural control and constrained optimization, the HNN needs to be appropriately designed to possess one (unique, singular) and globally stable  point of equilibrium, in order to mitigate the risk of spurious (and therefore incorrect and unwanted) responses that are bound to creep in due to the widespread problem of the presence of local minima in such circuits \cite{Gua00}. In order to accomplish these objectives, it has become customary to enforce certain restrictions (i.e. constraint conditions) on the weight matrix (or the interconnection matrix) of the network \cite{Ari00}. One pertinent example of such restrictions being imposed is as follows. The majority of study and applications of the Hopfield (and Hopfield-type) ANNs has been undertaken with the intrinsic presupposition that the interconnection (weight) matrix for the network under consideration is symmetric \cite{Xu96}.

This paper looks at the prospect of obtaining the solution of a system of $n$ simultaneous linear equations in $n$ variables using HNN-based approaches. Section~\ref{The Hopfield Network} contains an recapitulation of Hopfield's original proposal together with a deliberation on the reasons which lead to the standard HNN being not amenable for the purpose of solving sets of linear equations. The next section, i.e. Section~\ref{Modified Hopfield Network for solving Linear Equations} contains the circuit and explanation of a modified HNN being applied for obtaining the solution of  linear equations for the particular case wherein the coefficient matrix for the system of simultaneous linear equations is \textit{symmetric}. Computer simulation results using PSPICE software for different sets of linear equations (having symmetric weight matrices) are also included. Section~\ref{AHNN} highlights the application of a modified HNN for solving linear equations, in the case where the coefficient matrix  \textbf{A}, corresponding to the set of equations, is \textit{asymmetric}. Issues pertaining to the convergence and stability of the asymmetric HNN are also pondered upon. Section~\ref{sle-hnn:Circuit Simulation Results} presents the results of computer simulations for the proposed aysmmetric neural circuit employed for the solution of different sets of simultaneous linear equations of assorted sizes. Section~\ref{sle-hnn:Discussion} contains remarks related to the limitations, hardware complexity, applicability  and practicability of the proposed network. Section~\ref{sle-hnn:Conclusion} presents some concluding remarks.

\section{The Hopfield Network}
\label{The Hopfield Network}

\begin{figure} 
\begin{center}
\includegraphics[width=0.35\textwidth]{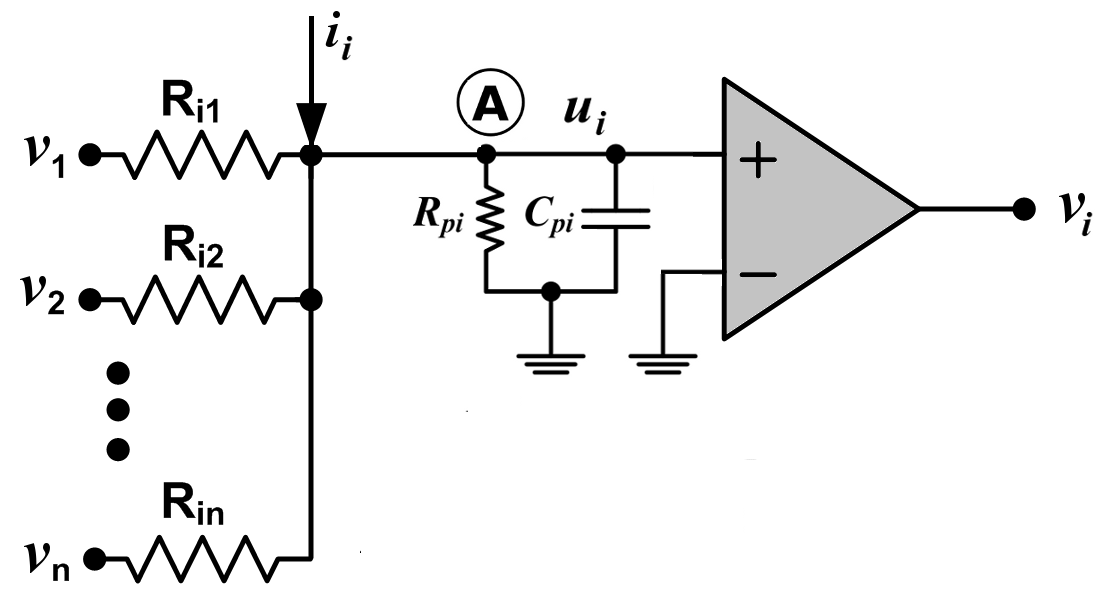}
\caption{$i$-th neuron in the standard HNN}
\label{fig:hopfield-ith-neuron}
\end{center}
\end{figure}

It is well know that the behaviour of a typical neuron (a genearlized $i$-th neuron in HNN is shown in Fig.~\ref{fig:hopfield-ith-neuron}) is modeled as a differential equation: 

\begin{displaymath}
C_i \frac{du_i}{dt} = \sum^n_{j=1} W_{ij}v_j - \frac{u_i}{R_i} + i_i
\end{displaymath}
\begin{equation}
v_i = g_i(u_i), i=1, 2, \ldots, n
\end{equation}
where $C_i$ $>$ 0, $R_i$ $>$ 0, and $i_i$ are the capacity (modeled electrically as capacitance), resistance value, and bias voltage applied to the neuron, respectively, and $u_i$ and $v_i$ are the input state and  the output state, respectively,  of the $i$-th neuron  and $g_i$(.) is the characteristic function defining the $i$-th neuron \cite{Hop85,JJH82,Hop84}. $W_{ij}$ are the individual elements, each having a different \{$i,j$\} combination, of the weight matrix \textbf{W} (which in turn decides the interconnections amongst the multiple neurons in the overall system). These weights are realized using appropriately valued resistors, $R_{ij} (= 1/W_{ij})$,  in the equivalent electronic (hardware) realization of the HNN, as has been illustrated in Fig,~\ref{fig:hopfield-ith-neuron}. It is to be noted that Hopfield performed his seminal study on such HNN networks with the following assumptions: 
\begin{equation} \label{eqn:hop-assum-1}
W_{ii} = 0~~ \textrm{for all $i$ values}
\end{equation}
and
\begin{equation} \label{eqn:hop-assum-2}
W_{ij} = W_{ji}~~ \textrm{for every $i,j$ combination}
\end{equation}

The assumptions mentioned above have the following implications on the hardware realization of the HNN.
\begin{itemize}
    \item Equation (\ref{eqn:hop-assum-1}) signifies that  self-interactions amongst the neurons (self-feedbacks in the hardware realization)  are not allowed  in the original version of the  HNN.
    \item Equation (\ref{eqn:hop-assum-2}) places a more rigorous condition on the allowable synergy between the  various units in the HNN in the sense that only symmetric connections (feedbacks) are allowed.  This leads to the restriction  that the HNN weight matrix \textbf{W} (=$(W_{ij})_{n\times{n}}$) has to be symmetric in nature.
\end{itemize}

It has indeed been proven that, for such type of cases, where self-interactions are disallowed (i.e. no self-feedbacks, because $W_{ii} =0$) and a symmetric \textbf{W} exists (i.e. $W_{ij}=W_{ji}$), the HNN is totally stable \cite{MVi93}. Under these restrictive conditions, the Lyapunov energy function defined for the HNN can be written as 
\begin{figure}
\includegraphics[width=0.48\textwidth]{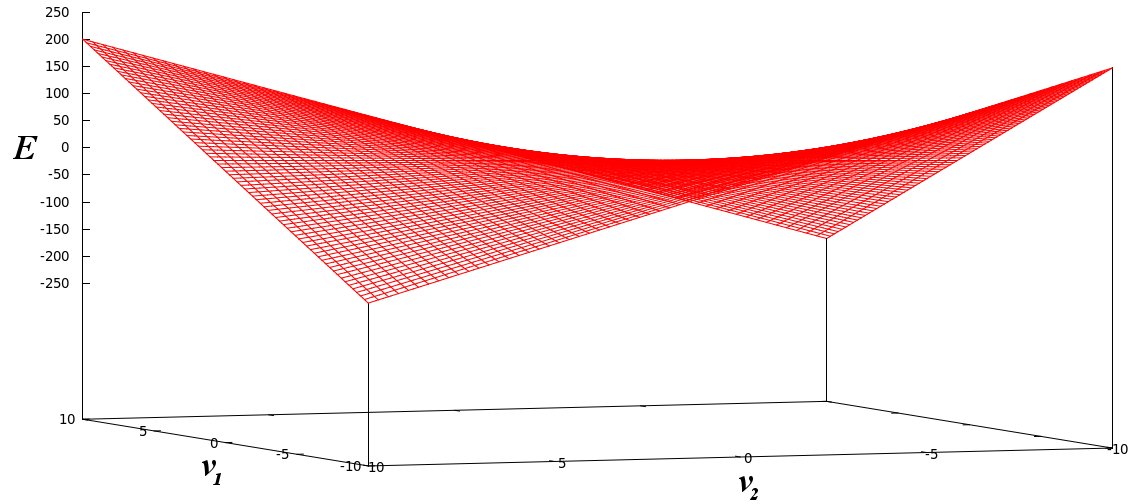}
\caption{Representative Lyapunov energy function plot for the standard HNN when applied with  2 neurons for the special scenario of a symmetric weight matrix \textbf{W} with all diagonal elements being equal to zero}
\label{fig:hopfield-ef-plot}
\end{figure}
\begin{equation} \label{eqn:ef-hopfield-for-ref}
E= - \frac{1}{2}\sum_i{\sum_j{W_{ij}v_i v_j}} - \sum_i{{i_iv}_i} + \sum_i \frac{1}{R_i} \int^{v_i}_0{{g_i}^{-1}}
\left({\mathbf v}\right)d{\mathbf v}
\end{equation}

The last term in (\ref{eqn:ef-hopfield-for-ref}) is usually neglected because of the following reason: it is only consequential near the saturating values of the operational amplifier used to emulate the neurons \cite{Tan86}. A 3D plot depicting the shape of a typical energy function, is shown in Figure~\ref{fig:hopfield-ef-plot}, for an HNN with the following specifications: 2--neuron system, symmetric weight matrix, all zeroes as diagonal elements in \textbf{W} (fulfilling the requirements laid out in (\ref{eqn:hop-assum-1}) and (\ref{eqn:hop-assum-2})), and no external bias $i_i$. It can readily be seen from Figure~\ref{fig:hopfield-ef-plot} that the stable states will be at the corner points of the virtual hypercube [-$V_m$,$V_m$] where ${\pm}V_m$ are the DC biasing supply voltages ($\pm$12-15$V$ for the $\mu$A-741 operational amplifier) of the opamps that are employed to realize the activation functions of the neurons.

Therefore, from the above discussion it is sufficiently clear that the standard HNN is unsuitable for the goal of solving linear equations, due to the points reproduced below:
\begin{itemize}
\item {Equation (\ref{eqn:hop-assum-1}) requires that \textbf{W} must always  have all zero diagonal elements}
\item {Equation (\ref{eqn:hop-assum-2}) dictates  that  \textbf{W} must always be symmetric in nature}
\item {The stable point(s) i.e. minima of the Lyapunov energy function $E$, obtained for the standard HNN, have been shown to occur at the corners of the hypercube, and it is not possible to shift these minima to lie at a particular point of interest, for instance, as is required if we want to use the HNN for solving linear equations -- we would want the minima to coincide with the mathematical solution point of the set of equations under consideration.}
\end{itemize}

\begin{figure}
\includegraphics[width=0.48\textwidth]{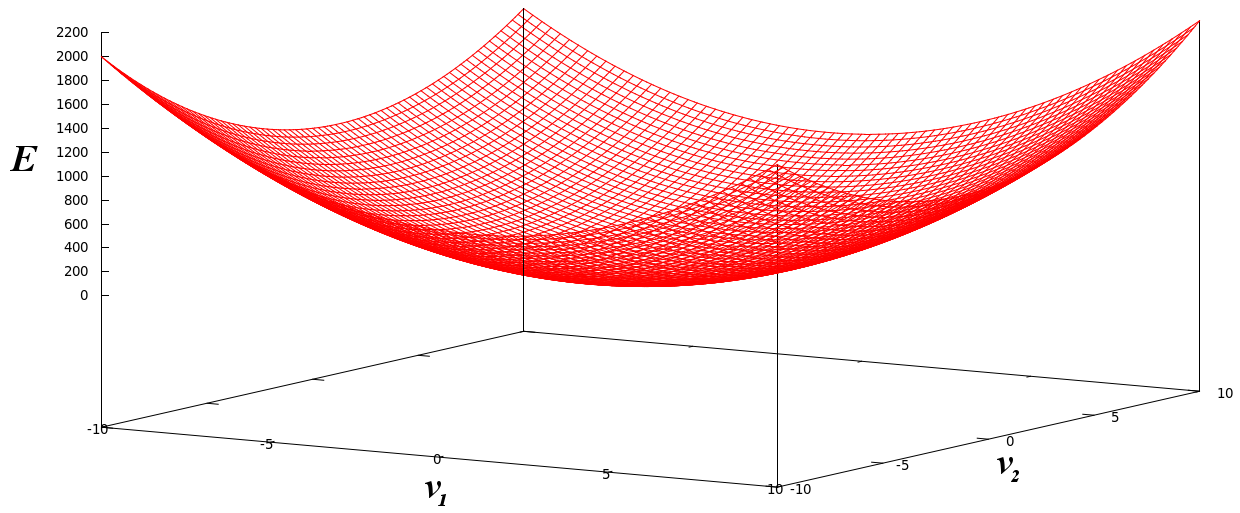}
\caption{representative Lyapunov energy function plotted for the modified HNN considering 2 neurons}
\label{fig:hopfield-ef-plot-mod}
\end{figure}

From the points noted above, it is quite evident that in order to make the standard HNN amenable for linear equation solving, some appropriate modifications would be required. One straight-forward modification, for instance, is the interchanging of the non-inverting and inverting inputs of the amplifiers in a standard HNN. The effect of this simple change is quite drastic on the energy function (\ref{eqn:ef-hopfield-for-ref}), and will cause it  to become:
\begin{equation} \label{eqn:ef-hopfield-mod-for-ref}
E= \frac{1}{2}\sum_i{\sum_j{W_{ij}v_i v_j}} + \sum_i{{i_iv}_i} - \sum_i \frac{1}{R_i} \int^{v_i}_0{{g_i}^{-1}}
\left({\mathbf v}\right)d{\mathbf v}
\end{equation}
The significance of the sign inversion in the energy function will become more evident when the above modification is coupled with the allowing of self-feedback (self-interactions) amongst the neurons \textit{i.e.} relaxation in the restrictive condition of (\ref{eqn:hop-assum-1}). These two modifications cause the minimum point of the Lyapunov energy function to now appear at the centre of the hypercube, and the same is presented in Figure~\ref{fig:hopfield-ef-plot-mod} for a pictorial comparison with the conventional HNN energy function shown in Fig.~\ref{fig:hopfield-ef-plot}.

Although an appropriately altered HNN, as has been discussed above, can be tailored to acquire a single (unique) minimum in the Lyapunov energy function, that stable minimum equilibrium  point can only exist at the centre of the hypercube \textit{i.e.} at $v_i = 0$, for all $i$. This leads to the following observation. The modified HNN is still not useful for being employed for the purpose of solving linear equations. This is because, for an ANN circuit to be able to do that, the minimum point in the Lyapunov energy function must coincide with the algebraic solution point of the set of linear equations. Such a solution point may be anywhere on the hyperplane, (and not compulsorily at the origin). The subsequent section considers some additional modifications/conditions to make the HNN applicable for the task under consideration in this paper.

\section{Solving Linear Equations using a Modified HNN}
\label{Modified Hopfield Network for solving Linear Equations}

The present section deals with  details of the modifications that are required in order to make the HNN (elaborated in the previous section) suitable for the purpose of providing the solution of different SLE. Let the system of $n$ SLE in $n$ variables to be solved be represented as 

\begin{equation} \label{eqn:single-layer-sle}
\mathbf {AV}= \mathbf B
\end{equation}
where
\begin{equation} \label{eqn:single-layer-sle-A}
{\mathbf A}=\ \left[ \begin{array}{cccc}
a_{11} & a_{12} & \dots  & a_{1n} \\ 
a_{21} & a_{22} & \dots  & a_{2n} \\ 
\vdots  & \vdots  & \dots  & \vdots  \\ 
a_{n1} & a_{n2} & \dots  & a_{nn} \end{array}
\right]
\end{equation}
\begin{equation} \label{eqn:single-layer-sle-B}
{\mathbf B}=\ \left[ \begin{array}{c}
b_1 \\ 
b_2 \\ 
\vdots  \\ 
b_n \end{array}
\right]
\end{equation}
\begin{equation} \label{eqn:single-layer-sle-V}
{\mathbf V}=\ \left[ \begin{array}{c}
V_1 \\ 
V_2 \\ 
\vdots  \\ 
V_n \end{array}
\right]
\end{equation}
where \textit{V${}_{1}$, V${}_{2}$, V${}_{3}$, \ldots , V${}_{n}$} are the unknowns and \textit{a${}_{ij}$} and \textit{b${}_{i}$} are coefficients and constants respectively, from the set of linear equations under consideration. Since a voltage-mode (VM) linear equation solver is presented in this work (meaning that the outputs of the circuit will be voltages equal in magnitude to the algebraic solution), the decision variables are labelled as voltages \textit{V${}_{1}$, V${}_{2}$, V${}_{3}$, \ldots , V${}_{n}$} in order to match the output states of neurons in the proposed circuit. The proposed network is based on the assumption that the coefficient matrix \textbf{A} is invertible (nondegenerate). This means that the system of LE (\ref{eqn:single-layer-sle}) is not under-determined and is consistent. This also means that the linear system (\ref{eqn:single-layer-sle}) has a uniquely determined solution. Moreover, there is another restriction that needs to be imposed considering the electronic hardware implementation of the HNN. The solution must not fall beyond the operational region (in terms of voltages) of the circuit \textit{i.e.} inside the hypercube defined by $|V_i|$ $\leq$ $V_m$ ($i$ = 1, 2, 3, \ldots, $n$). This restriction is due to the fact that the output obtained from an operational amplifier (which is used to emulate the neuron) cannot go beyond its biasing voltage.

\begin{figure} 
\includegraphics[width=0.42\textwidth]{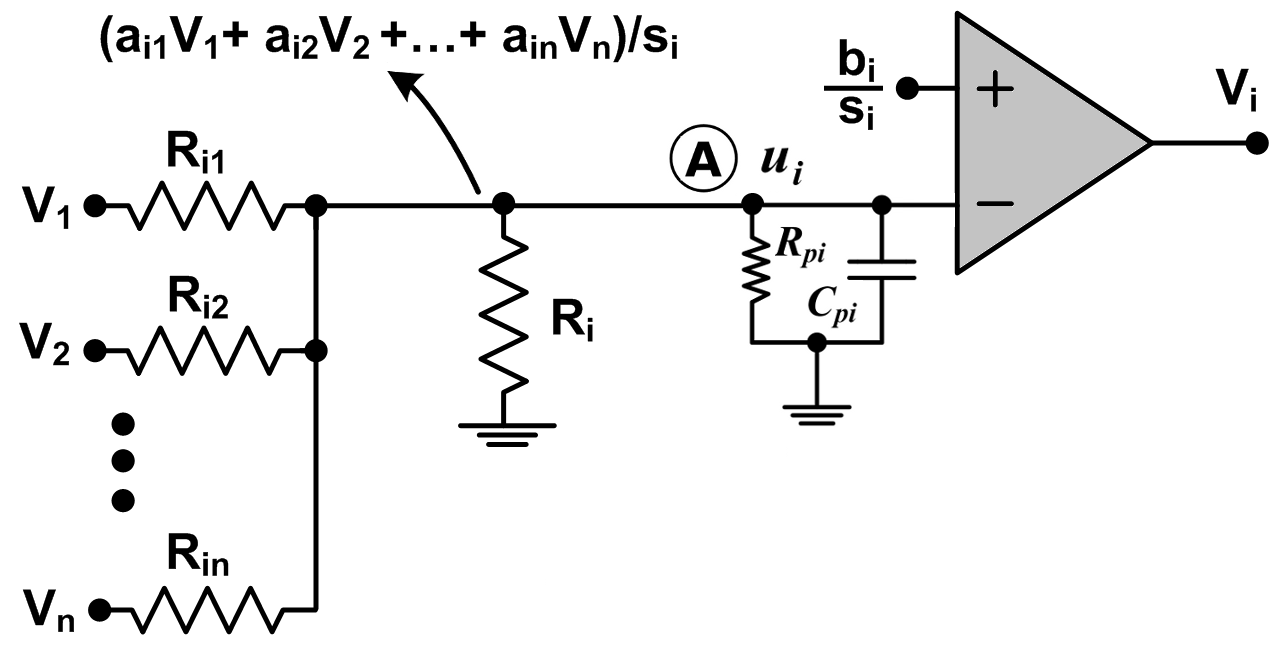}
\caption{$i$-th neuron in the modified HNN applied for obtaining the  solution of linear equations}
\label{fig:sle-single-layer-generic}
\end{figure}

The generalized $i$-th neuron of the VM modified HNN based circuit for solving  (\ref{eqn:single-layer-sle}) is depicted in Figure~\ref{fig:sle-single-layer-generic}. It needs to be mentioned that the output voltage of the generalized neuronal amplifier is now labelled as $V_i$ (in place of $v_i$ in Figure~\ref{fig:hopfield-ith-neuron}) to comply with the notation used for the unknowns i.e. the decision variables in (\ref{eqn:single-layer-sle-V}).  \textit{C${}_{pi}$} and  \textit{R${}_{pi}$}  are the parasitic capacitance and parasitic resistance respectively,   of the voltage amplifier emulating the \textit{i}-th neuron. These circuit parasitics  have been considered for modelling the time-varying (dynamic) nature of the output characteristics of the voltage amplifier. As is evident from Figure~\ref{fig:sle-single-layer-generic}, individual equations from the SLE (\ref{eqn:single-layer-sle}) have been scaled (by a factor $s_i$) prior to their application as inputs to the HNN neurons. This scaling is essential for ensuring that all \textbf{[$a_{ij}/s_i$]} coefficients are less than unity, which in turn means that these could be realized in hardware with passive voltage dividers implemented with discrete resistors. As will be elaborated subsequently , the magnitudes of the individual scaling factors ($s_i$) could be selected independently for each equation. If that is the case, the scaling factor for the $i$-th LE, $s_i$, should adhere to:
\begin{equation} \label{eqn:sle-single-layer-si}
s_i ~ \ge ~ \sum_{j=1}^n a_{ij}
\end{equation}

All equations in the set to be solved could also be scaled by the same magnitude $s$, which  should be the highest amongst all $s_i$ selected for individual LE separately \textit{i.e.}
\begin{equation} \label{eqn:sle-single-layer-s}
s = \textrm{max} (s_i); ~~~ \textrm{for all}~i
\end{equation}

Considering the circuit of Figure~\ref{fig:sle-single-layer-generic}, and writing the node equation for the node marked `A' (i.e.  the inverting input of the operational amplifier) yields the equation of motion for  \textit{i}-th neuron in the following form:
\begin{equation} \label{eqn:single-layer-sle-ith-current}
C_{pi}\frac{du_i}{dt}{=\ }\frac{V_{{1}}}{R_{{i}1}}{+\ }\frac{V_{{
 2}}}{R_{{ i}2}}{ +\dots +\ }\frac{V_n}{R_{in}}{ -}u_i\left[\frac{{
1}}{R_{eqv,i}}\right]
\end{equation}
where \textit{u${}_{i}$} represents the internal state in the \textit{i}-th neuron, and
\begin{equation} \label{eqn:single-layer-sle-Ri}
R_{eqv,i} = R_{i1} \left\|R_{i2} \right\| R_{in} \ldots \left\| R_i \right\| R_{pi}
\end{equation}

For obtaining a valid Lyapunov energy function for the modified HNN depicted in Figure~\ref{fig:sle-single-layer-generic}, we should proceed in the following manner. The association between $V_i$ and $u_i$ for the opamp (emulating the neuron) in Figure~\ref{fig:sle-single-layer-generic} is given as
\begin{equation} \label{eqn:ui-vi-relationship-1}
V_i = g_i \left( \frac{b_i}{s_i} - u_i \right)
\end{equation}
which can be rearranged to the following form
\begin{equation} \label{eqn:ui-vi-relationship-2}
u_i = \frac{b_i}{s_i} - {g_i}^{-1}(V_i)
\end{equation}

Further, for dynamical systems such as  the one depicted in Figure~\ref{fig:sle-single-layer-generic}, the gradient of the Lyapunov energy function $E$ is concomitant to the dynamical progression-in-time  of the circuit, as represented by (\ref{eqn:ef-hopfield-de-dvi}) \cite{Zur92}.
\begin{equation} \label{eqn:ef-hopfield-de-dvi}
\frac{\partial E}{\partial V_i} =  C_{pi} \frac{du_i}{dt}; ~~~ \textrm{for all} ~ i
\end{equation}

Using (\ref{eqn:single-layer-sle-ith-current}, \ref{eqn:ef-hopfield-de-dvi}), along with (\ref{eqn:hop-assum-2}), the energy function $E$ corresponding to the modified HNN circuit of Figure~\ref{fig:sle-single-layer-generic} can be written as
\begin{equation} \nonumber
E= \frac{1}{2}\sum_i{\sum_j{W_{ij}V_i V_j}} - \sum_i \frac{b_i/s_i}{R_{eqv,i}} V_i 
\end{equation}
\begin{equation} \label{eqn:ef-sle-single-layer}
+ \sum_i \frac{1}{R_{eqv,i}} \int^{V_i}_0{{g_i}^{-1}}
\left({\mathbf V}\right)d{\mathbf V}
\end{equation}

The last term in (\ref{eqn:ef-sle-single-layer}) is insignificant when compared with the first two terms, and is therefore usually neglected in further analyses \cite{Tan86, Zur92}. Thus, the Lyapunov energy function may further be written in a reduced and simplified form as:
\begin{equation} \label{eqn:ef-sle-single-layer-simple}
E= \frac{1}{2}\sum_i{\sum_j{W_{ij}V_i V_j}} - \sum_i \frac{b_i/s_i}{R_{eqv,i}} V_i
\end{equation}

From (\ref{eqn:ef-sle-single-layer-simple}), it is clearly evident that the minimum (and therefore the most stable point, to which the circuit would converge) of $E$ will also be dictated by the individual constants contained in the vector \textbf{B} in (\ref{eqn:single-layer-sle-B}). This goes on to show that the minimum point will now not (always) appear at the centre of the $n$-dimensional hypercube, which is in contrast to the energy function derived in (\ref{eqn:ef-hopfield-mod-for-ref}) which necessitated the minima to lie at the origin.

The stationary point (minima) of $E$  (\ref{eqn:ef-sle-single-layer-simple}) may be obtained by setting
\begin{equation} \label{eqn:ef-sle-single-layer-simple-st-pt-1}
\frac{\partial E}{\partial V_i} =  0;   ~~~ i = 1, 2, 3, \ldots, n
\end{equation}
from where, we get
\begin{equation} \label{eqn:ef-sle-single-layer-simple-st-pt-2}
\frac{V_{{1}}}{R_{{ i}1}}{ +\ }\frac{V_{{ 2}}}{R_{{ i}2}}{ +\dots +\ }\frac{V_n}{R_{in}} 
- \frac{b_i/s_i}{R_{eqv,i}} = 0; ~~~ i = 1, 2, 3, \ldots, n
\end{equation}

The set of linear equations of (\ref{eqn:single-layer-sle}), after scaling, can be written in a generalized form as
\begin{equation} \label{eqn:ef-sle-single-layer-simple-st-pt-3}
\frac {a_{i1} V_1}{s_i} + \frac {a_{i2} V_2}{s_i} + \ldots + \frac {a_{in} V_n}{s_i} - \frac{b_i}{s_i} = 0; ~~~ i = 1, 2, 3, \ldots, n
\end{equation}

For the equilibrium point (minima) of $E$ given in  (\ref{eqn:ef-sle-single-layer-simple}), which was derived in (\ref{eqn:ef-sle-single-layer-simple-st-pt-2}), to overlap with the solution  of the SLE (\ref{eqn:ef-sle-single-layer-simple-st-pt-3}), the magnitudes of the resistors in the circuit may be obtained as follows: 
\begin{equation} \label{eqn:ef-sle-single-layer-rij-values}
R_{ij} = \frac {s_i}{a_{ij}}; ~~~  i, j = 1, 2, 3, \ldots, n 
\end{equation}

The value of the individual resistors $R_i$, may be computed using (\ref{eqn:ef-sle-single-layer-simple-st-pt-2},  \ref{eqn:ef-sle-single-layer-simple-st-pt-3}) by equating their respective last terms, and resulting in
\begin{equation}
\frac{1}{R_{eqv,i}} = 1  
\end{equation}
which can be expanded to
\begin{equation} \label{eqn:ef-sle-single-layer-ri-values-intermediate}
\frac{1}{R_{i1}} + \frac{1}{R_{i2}} +\ldots + \frac{1}{R_{in}} + \frac{1}{R_i} + \frac{1}{R_{pi}} = 1  
\end{equation}

Also, since the value of $R_{pi}$ is much greater than all  other resistances appearing in (\ref{eqn:ef-sle-single-layer-ri-values-intermediate}), $R_{pi}$ can readily be neglected when the equivalent value of all the resistors (which appear in parallel configuration) connected to the $i$-th input node (inverting terminal of the $i$-th opamp) is computed. Therefore, we have
\begin{equation} \label{eqn:ef-sle-single-layer-ri-values-intermediate-2}
\frac{1}{R_{i1}} + \frac{1}{R_{i2}} +\ldots + \frac{1}{R_{in}} + \frac{1}{R_i} = 1  
\end{equation}

Putting the values of resistances $R_{ij}$ obtained from (\ref{eqn:ef-sle-single-layer-rij-values}) into (\ref{eqn:ef-sle-single-layer-ri-values-intermediate-2}) yields
\begin{equation} \label{eqn:ef-sle-single-layer-ri-values}
R_i  =  \frac{s_i}{s_i - \sum_{j=1}^n a_{ij}}  
\end{equation}

It is evident that, the constraint being enforced on the individual scaling factors (in (\ref{eqn:sle-single-layer-si})) can also be derived  from (\ref{eqn:ef-sle-single-layer-ri-values}), because since selecting a vlaue of scaling factor which violates (\ref{eqn:sle-single-layer-si}) is bound to result in a \textit{negative} resistor value from (\ref{eqn:ef-sle-single-layer-ri-values}).

Referring to (\ref{eqn:ef-sle-single-layer-rij-values}), the magnitudes of  the weight resistors in the modified HNN when used to solve SLE can be computed as
\begin{equation} \label{eqn:sle-hopfield-mod-rij-values}
\left[ \begin{array}{cccc}
R_{11}  & R_{12}  & \dots   & R_{1n} \\ 
R_{21}  & R_{22}  & \dots   & R_{2n} \\ 
\vdots  & \vdots  & \ddots  & \vdots  \\ 
R_{n1}  & R_{n2}  & \dots   & R_{nn} \end{array}
\right]
=
\left[ \begin{array}{cccc}
\frac {1} {a_{11}/s_1} & \frac {1} {a_{12}/s_1} & \dots  & \frac {1} {a_{1n}/s_1}   \\ 
\frac {1} {a_{21}/s_2} & \frac {1} {a_{22}/s_2} & \dots  & \frac {1} {a_{2n}/s_2}   \\ 
\vdots                 & \vdots                 & \ddots & \vdots                 \\ 
\frac {1} {a_{n1}/s_n} & \frac {1} {a_{n2}/s_n} & \dots  & \frac {1} {a_{nn}/s_n} \end{array}
\right]
\end{equation}

\begin{figure} 
\includegraphics[width=0.48\textwidth]{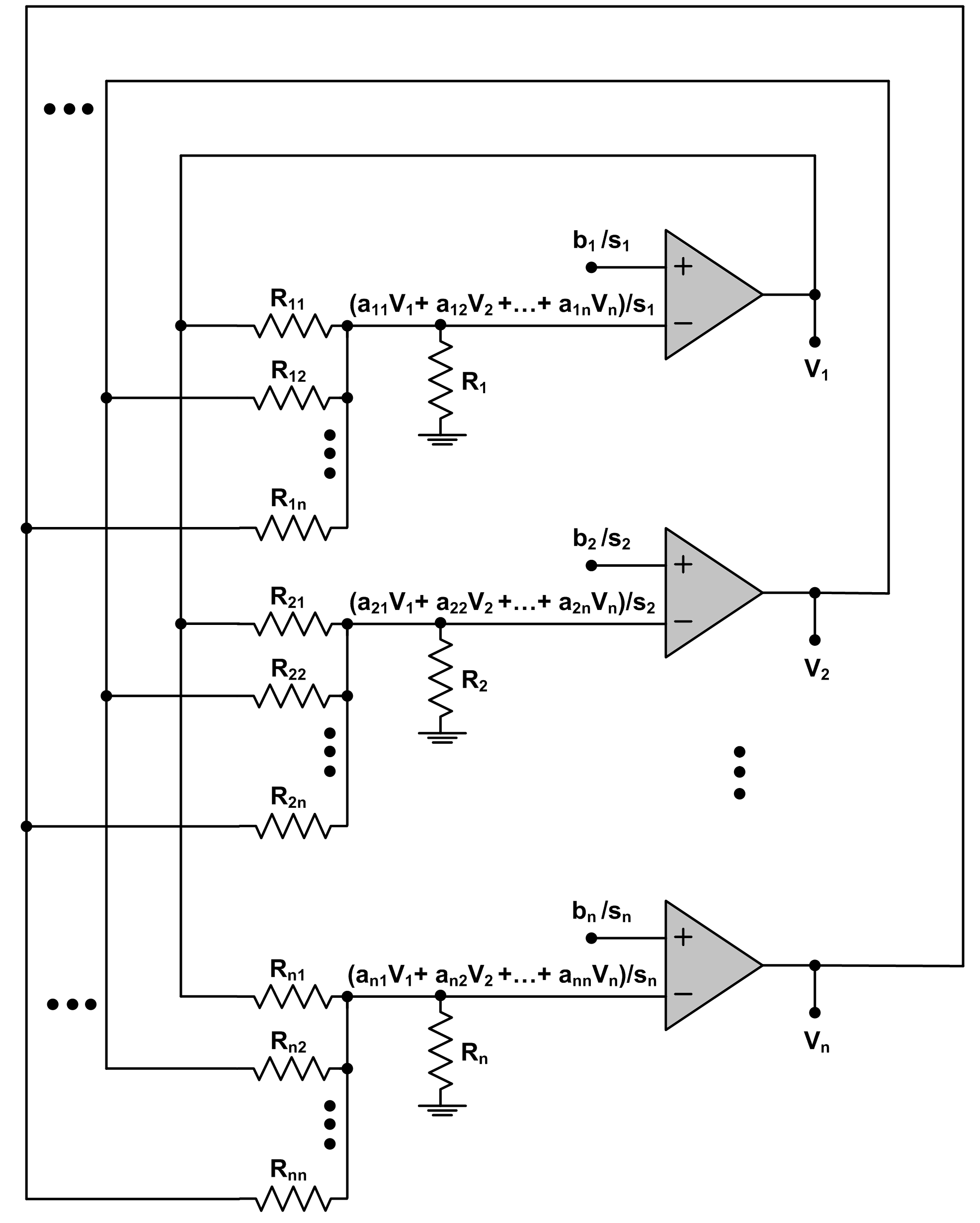}
\caption{Complete circuit of the modified HNN, as applied for the objective of this work}
\label{fig:sle-single-layer-full-network}
\end{figure}

The full generalized version of the \textit{symmetric} HNN employed  for solving  $n$  equations in $n$ variables obtained by using the neurons rendered in Figure~\ref{fig:sle-single-layer-generic} is illustrated in Figure~\ref{fig:sle-single-layer-full-network}. 

For verifying  that the modified HNN-based  equation solving circuit of Figure~\ref{fig:sle-single-layer-full-network} is indeed capable of converging to solutions of SLE, the network was tested using PSPICE simulations. The restriction of the   coefficient matrix (\textbf{A}) being necessarily symmetric is always considered while generating test sets of SLE. First, an example  set of 2  equations with 2 variables (\ref{eqn:sle-hopfield-2-var}) was solved using the circuit in Figure~\ref{fig:sle-single-layer-full-network}. Values of  resistors (in K$\Omega$)  obtained using (\ref{eqn:sle-hopfield-mod-rij-values}) and (\ref{eqn:ef-sle-single-layer-ri-values}) for the example set of  equations (\ref{eqn:sle-hopfield-2-var}) have been listed below. Both equations were scaled by $s_1=s_2$=10.

\begin{figure}
\begin{center}
\includegraphics[width=0.28\textwidth]{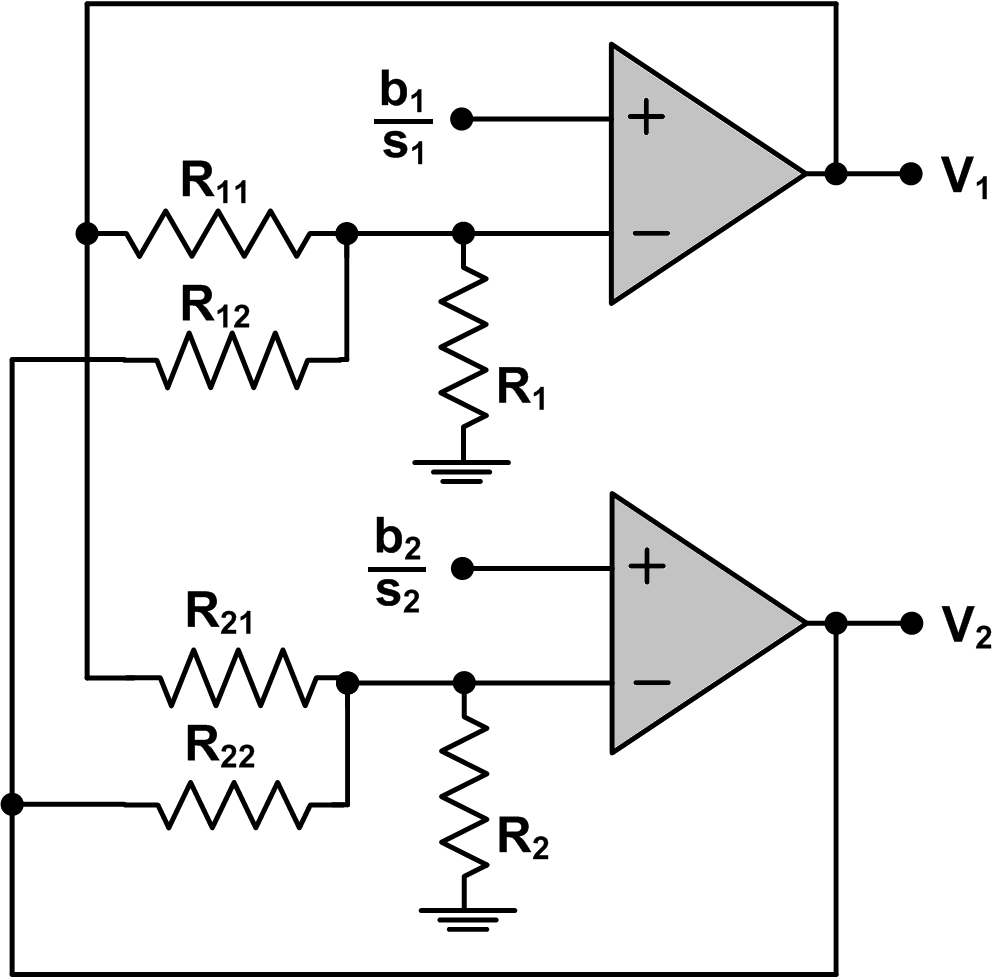}
\caption{The modified HNN,  as applied for solving 2 simultaneous linear equations, with the restriction of a symmetric \textbf{A} matrix}
\label{fig:sle-single-layer-2-var}
\end{center}
\end{figure}
\begin{figure}
\begin{center}
\includegraphics[width=0.48\textwidth]{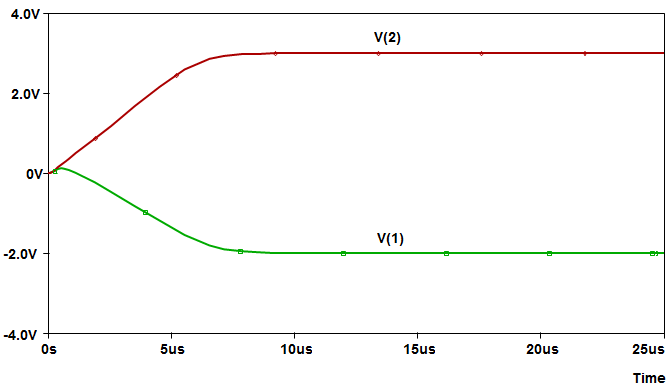}
\caption{Computer simulation results using PSPICE, for the circuit of Figure~\ref{fig:sle-single-layer-2-var} employed to obtain solution for (\ref{eqn:sle-hopfield-2-var})}
\label{fig:sle-hopfield-2-var-result}
\end{center}
\end{figure}
\begin{equation} \label{eqn:sle-hopfield-2-var}
\left[ 
\begin{array}{cc}
4  &  3 \\ 
3  &  5  
\end{array}
\right]
\left[ 
\begin{array}{c}
V_1 \\ 
V_2 
\end{array}
\right]
=
\left[ 
\begin{array}{c}
1 \\ 
9 
\end{array}
\right]
\end{equation}
\begin{equation} \label{eqn:sle-hopfield-2-var-rij}
\left[ 
\begin{array}{cc}
R_{11}  &  R_{12} \\ 
R_{21}  &  R_{22}  
\end{array}
\right]
=
\left[ 
\begin{array}{cc}
2.5   &  3.33  \\ 
3.33   &  2   
\end{array}
\right]
\end{equation}
and
\begin{equation} \label{eqn:sle-hopfield-2-var-ri}
\left[ 
\begin{array}{c}
R_1  \\ 
R_2 
\end{array}
\right]
=
\left[ 
\begin{array}{c}
3.33 ~K{\Omega}  \\ 
5 ~K{\Omega}  
\end{array}
\right]
\end{equation}

The circuit to solve the 2 SLE in 2 variables (\ref{eqn:sle-hopfield-2-var}), drawn using the generalized network of Figure~\ref{fig:sle-single-layer-full-network}, is illustrated in Figure~\ref{fig:sle-single-layer-2-var}. It can be readily found that the mathematical solution for (\ref{eqn:sle-hopfield-2-var}) is $V_1$ = $-$2 and $V_2$ = 3. Results of computer simulation (using PSPICE software)  using the resistor values provided in (\ref{eqn:sle-hopfield-2-var-rij}, \ref{eqn:sle-hopfield-2-var-ri}) are included in Figure~\ref{fig:sle-hopfield-2-var-result}. It can be seen that the resultant output node voltages (measured in Volts) converge to V(1) = $-$2.00 and V(2) = 3.00, which correspond perfectly with the mathematical solution.

\section{Solving SLE with Asymmetric HNN (AHNN)}
\label{AHNN}

The preceding section dealt with a discussion on how the standard HNN may be appropriately modified for the solution of systems of $n$ SLE in $n$ variables, although with the strict restriction that the coefficient matrix \textbf{A} is symmetric. This assumption, leading to \textbf{W} being symmetric, formed the bedrock of the development of the theory of stability of the HNN, as well as the presence of a valid/legitimate Lyapunov energy function. Customarily, weights are realized using resistances in the conventional electronic implementations of the HNN, and therefore the strict condition of symmetry of \textbf{W} dictates that meticulously controlled resistor values be used in the circuit realizations for preserving and maintaining the symmetry of \textbf{W}.  However, when the HNN is employed for real-world (practical) scenarios using appropriate hardware implementations, it is impractical to expect, and impossible to guarantee, that all the neuronal interactions are symmetric in nature, since this will require ensuring that two electrical quantities (i.e. values of resistances, voltage gains of the opamps used as neuronal amplifiers) are of the exact same value.  Therefore, a physically implemented HNN circuit, would in all probability, actually turn out to be one with an \textit{asymmetric} weight matrix. Furthermore, as Vidyasagar has aptly  pointed out in his work \cite{MVi93}:

\textit{``The consequences of even slight asymmetries in the interactions are disastrous to the theory of stability as put forward by Hopfield"} 

\noindent a substantial quantum of research endeavor has already been targeted towards the study and understanding of the qualitative properties of oscillation, convergence, and stability for AHNN. \cite{Xu96,Gua00,MVi93,Che01,Ari00}. 

In the existing technical literature, the sufficient conditions which can guarantee that an AHNN has a unique (exponentially) stable equilibrium state, have been presented in several different manifestations:

\begin{itemize}
\item {diagonally row or column dominance property in \textbf{W} \cite{Ari00}}
\item {bounded-ness of the activation functions of the neurons \cite{Lia98}}
\item {some restrictions on interconnection matrix of the AHNN \cite{Yan94}}
\item {negative semi-definiteness of a matrix derived from the interconnection matrix of the AHNN \cite{For94}}
\item {\textit{M}-matrix characteristics being exhibited by a matrix derived from \textbf{W} \cite{Mic89}}
\item {diagonal stability \cite{Kas94}}
\end{itemize}

In addition to the above, the presence or absence of a unique equilibrium and the consequent stability of an AHNN can also be ascertained with a new collection of simplified sufficient conditions \cite{Gua00}. The condition for which a normalized AHNN is stable, as has been shown mathematically in \cite{Gua00}. The same has been reproduced here for ready reference of the readers. As a precondition, first the \textit{M}-matrix is defined as follows.

\textit{Let \textbf{T} (=$(t_{ij})_{n\times{n}}$) be a square matrix which has non-positive off-diagonal elements. Now, if each of  the leading principal minors of \textbf{T} is positive, the matrix \textbf{T} is referred to as an \textit{M}-matrix  \cite{Gua00}. }

The following steps are therefore required to determine whether an AHNN has a unique equilibrium and is stable:
\begin{enumerate}
    \item A square matrix \textbf{T} (=$(t_{ij})_{n\times{n}}$) is first obtained from \textbf{W} using (\ref{eqn:calc-a-matrix}).
    \item If \textbf{T} is a valid \textit{M}-matrix, then the AHNN pertaining to  \textbf{W} is declared to be globally asymptotically stable \cite{Gua00}.
\end{enumerate}
, 
\begin{equation} \label{eqn:calc-a-matrix}
t_{ij} = \left\{ \begin{array}{cc}
1-W_{ij}~; & i=j   \\ 
-\left | W_{ij} \right |~; & i\neq j
\end{array}\right.
\end{equation}

It is therefore clear that (\ref{eqn:calc-a-matrix})  puts some  `restrictions' on  \textbf{W}. This can be understood as follows. An AHNN will not necessarily possess a unique equilibrium for all \textbf{W}. Instead, only those interconnection matrices \textbf{W}, from which the counterpart \textbf{T} matrix obtained using (\ref{eqn:calc-a-matrix}), turns out to be a valid \textit{M}-matrix, are admissible. This restriction on  \textbf{W} leads to a limitation in the application of AHNN to real-world problems.  This is also true for the linear equation solver circuit presented here, where  it is observed that the proposed network is capable of solving only those  SLE for which \textbf{W} satisfies the conditions discussed above. 

In the succeeding section, the application of an AHNN for the solution of sets of SLE of varying sizes is presented. It should be mentioned  at this juncture that although the prospect of the application  of AHNN for the  solution of  SLE has always been in existance since the advent of the standard HNN itself, a thorough search of the technical literature does \textit{not} reveal any such attempt. To the best of the author's knowledge, the work contained in this paper is the first pursuit to exploit AHNNs for the purpose of obtaining solutions to  systems of SLE.

\section{Circuit Simulation Results}
\label{sle-hnn:Circuit Simulation Results}

The correct operation of the AHNN used for solving linear equations was verified with extensive PSPICE simulations. Different  sets of linear equations, starting with 2 and going up to 20 variables, were generated, and solved by the proposed AHNN circuit. Results obtained from the computer simulations are presented in this section, and were found to be in close agreement with the mathematically obtained  solutions, for all the equation sets tested.

The circuit of Figure~\ref{fig:sle-single-layer-2-var} was used to obtain the solution of the following simultaneous linear equations in 2 variables:

\begin{equation} \label{eqn:sle-single-layer-2-var}
\left[ 
\begin{array}{cc}
3  &  2 \\ 
7  &  8  
\end{array}
\right]
\left[ 
\begin{array}{c}
V_1 \\ 
V_2 
\end{array}
\right]
=
\left[ 
\begin{array}{c}
-1 \\ 
-9 
\end{array}
\right]
\end{equation}

The values of the resistors (in $K{\Omega}$) as calculated from (\ref{eqn:ef-sle-single-layer-rij-values}) and (\ref{eqn:ef-sle-single-layer-ri-values}) for the example set of 2 linear equations (\ref{eqn:sle-single-layer-2-var}) are presented in (\ref{eqn:sle-single-layer-2-var-rij}, \ref{eqn:sle-single-layer-2-var-ri}). The magnitudes of the scaling factors for the two equations were chosen to be $s_1$ = 6 and $s_2$ = 21 respectively. It should be noted that the scaling factors are to be chosen such that the resulting values of the resistor(s) are realistic and reasonable. For instance, choosing a very low value of the scaling factor ($< 1$) would result in a very low value of the resistances when the coefficients in the linear equations are positive integers higher than unity. Similarly, selection of a very high value of the scaling factor(s) would result in high values of the resistors, which is inadvisable from an integrated circuit fabrication viewpoint.

\begin{equation} \label{eqn:sle-single-layer-2-var-rij}
\left[ 
\begin{array}{cc}
R_{11}  &  R_{12} \\ 
R_{21}  &  R_{22}  
\end{array}
\right]
=
\left[ 
\begin{array}{cc}
2   &  3  \\ 
3   &  2.625   
\end{array}
\right]
\end{equation}
and
\begin{equation} \label{eqn:sle-single-layer-2-var-ri}
\left[ 
\begin{array}{c}
R_1  \\ 
R_2 
\end{array}
\right]
=
\left[ 
\begin{array}{c}
6   \\ 
3.5   
\end{array}
\right]
\end{equation}
\begin{figure}
\includegraphics[width=0.48\textwidth]{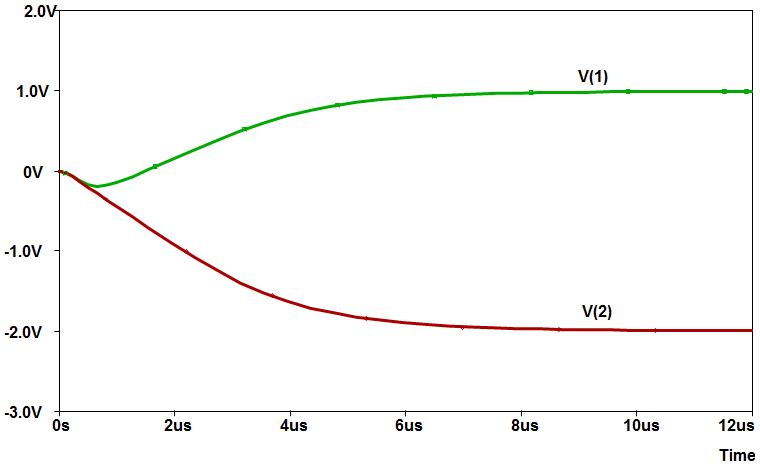}
\caption{Results of computer simulations for the AHNN applied to solve (\ref{eqn:sle-single-layer-2-var})}
\label{fig:sle-single-layer-2-var-result}
\end{figure}

Results of the PSPICE simulation test for the circuit of Figure~\ref{fig:sle-single-layer-2-var}, which is used to solve (\ref{eqn:sle-single-layer-2-var}) using the resistor values shown in (\ref{eqn:sle-single-layer-2-var-rij}) and (\ref{eqn:sle-single-layer-2-var-ri}) are depicted in Figure~\ref{fig:sle-single-layer-2-var-result} from where it can be observed that the obtained (stable, settled)  voltages at the output nodes are V(1,2) = (1.01, $-$2.01).  These match closely with the mathematical solution: $V_1$ = 1 and $V_2$ = $-$2. Thereafter, different sets of SLE in 2 variables were solved utilizing the circuit of the AHNN. The PSPICE simulations results are tabulated in Table~\ref{tab:sle-single-layer-sim-ver-results-1}. It can be readily observed from Table~\ref{tab:sle-single-layer-sim-ver-results-1} that the AHNN can provide correct solutions for all the chosen sets of SLE for which the criterion of stability (discussed in Section~\ref{AHNN}) is fulfilled. It is important to note that each of the matrix \textbf{T}  shown in Table~\ref{tab:sle-single-layer-sim-ver-results-1} (for the different cases) is calculated by utilizing $-$\textbf{A} (instead of \textbf{A}). This is required since the criterion of stability  presented in Section~\ref{AHNN} is mathematically obtained for the case when the amplifier emulating the neuron in the AHNN has a monotonically increasing activation function,  whereas for the case of the circuit of the AHNN that is actually applied to solve the different sets of SLE, the opamp which is used in the inverting configuration, exhibits  monotonically decreasing transfer characteristics.

A system of 3 linear equations in 3 variables  (\ref{eqn:sle-single-layer-3-var}) was next solved using the proposed AHNN. Values of the different resistors required in the AHNN, as calculated using (\ref{eqn:ef-sle-single-layer-rij-values}) and (\ref{eqn:ef-sle-single-layer-ri-values}) for the chosen set of 3 SLE in 3 variables (\ref{eqn:sle-single-layer-3-var}) are presented below. All the three LE were scaled by $s_1$ = $s_2$ = $s_3$ = 10.

\begin{equation} \label{eqn:sle-single-layer-3-var}
\left[ 
\begin{array}{ccc}
4  &  2  &  1 \\ 
2  &  7  &  1 \\ 
3  &  1  &  6
\end{array}
\right]
\left[ 
\begin{array}{c}
V_1 \\ 
V_2 \\
V_3
\end{array}
\right]
=
\left[ 
\begin{array}{c}
2 \\ 
-10 \\
13
\end{array}
\right]
\end{equation}
\begin{equation} \label{eqn:sle-single-layer-3-var-rij}
\left[ 
\begin{array}{ccc}
R_{11}  &  R_{12}  &  R_{13}  \\ 
R_{21}  &  R_{22}  &  R_{23}  \\
R_{31}  &  R_{32}  &  R_{33}  
\end{array}
\right]
=
\left[ 
\begin{array}{rrr}
2.5 ~K{\Omega}   &  5 ~K{\Omega}      &  10 ~K{\Omega}  \\ 
5 ~K{\Omega}     &  1.428 ~K{\Omega}  &  10 ~K{\Omega}  \\ 
3.33 ~K{\Omega}  &  10 ~K{\Omega}     &  1.666 ~K{\Omega}
\end{array}
\right]
\end{equation}
and
\begin{equation} \label{eqn:sle-single-layer-3-var-ri}
\left[ 
\begin{array}{c}
R_1  \\ 
R_2  \\
R_3
\end{array}
\right]
=
\left[ 
\begin{array}{c}
3.33 ~K{\Omega}  \\ 
\infty  \\ 
\infty  
\end{array}
\right]
\end{equation}
\begin{figure}
\includegraphics[width=0.48\textwidth]{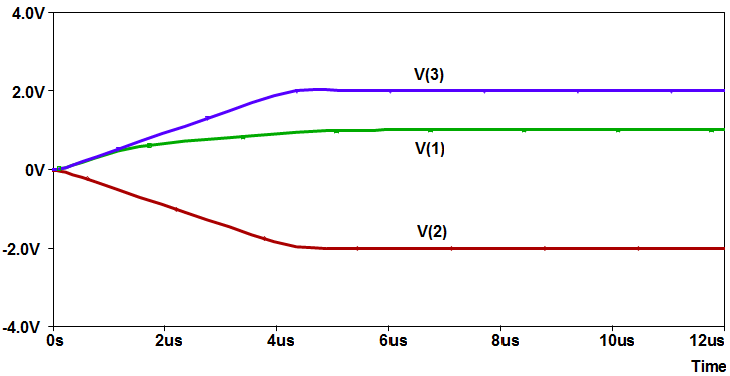}
\caption{Results of computer simulations for the AHNN used to solve (\ref{eqn:sle-single-layer-3-var})}
\label{fig:sle-single-layer-3-var-result}
\end{figure}
\begin{figure}
\includegraphics[width=0.48\textwidth]{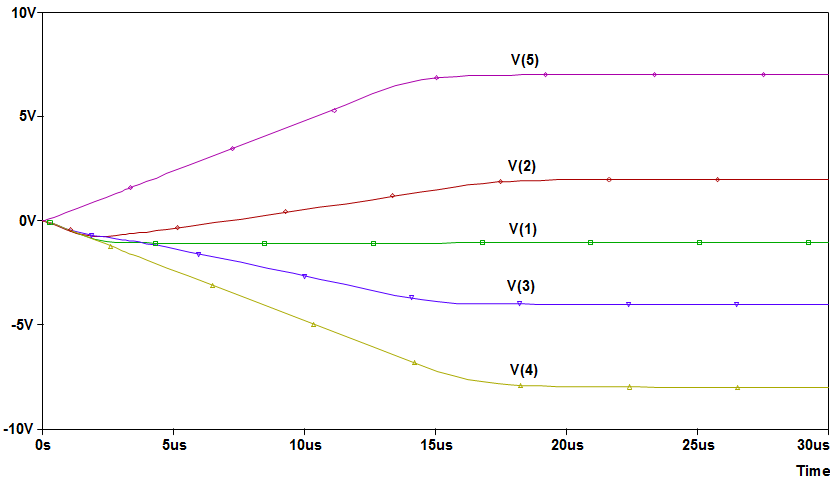}
\caption{Results of computer simulations for the AHNN employed for obtaining the correct answers for the first set of  equations in Table~\ref{tab:sle-single-layer-sim-ver-results-3}}
\label{fig:sle-single-layer-5-var-result}
\end{figure}
\begin{figure}
\includegraphics[width=0.48\textwidth]{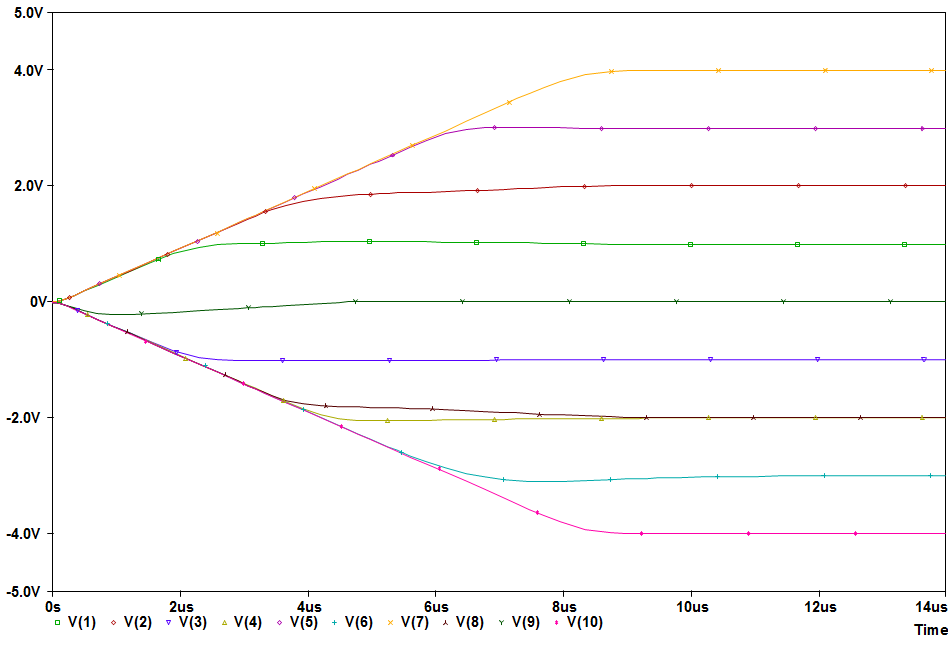}
\caption{Results as obtained after simulations, for the AHNN being used to solve the 10-variables set of  equations listed in Table~\ref{tab:sle-single-layer-large-results}}
\label{fig:sle-single-layer-10-var-result}
\end{figure}
\begin{figure}
\includegraphics[width=0.48\textwidth]{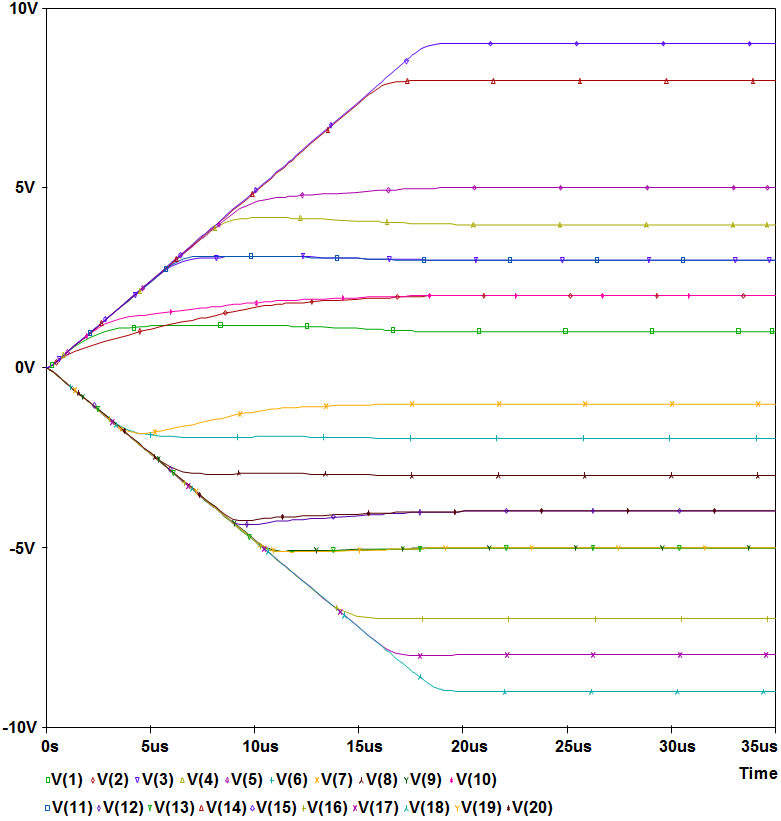}
\caption{Results that were obtained after simulation, for the AHNN put to solve the 20-variables linear equation system listed in Table~\ref{tab:sle-single-layer-large-results}}
\label{fig:sle-single-layer-20-var-result}
\end{figure}
\begin{table}
\caption{Results of the AHNN used to converge to solutions of various sets of linear equations in 2 variables}
\label{tab:sle-single-layer-sim-ver-results-1}
\begin{tabular}{c}
\includegraphics[width=0.48\textwidth]{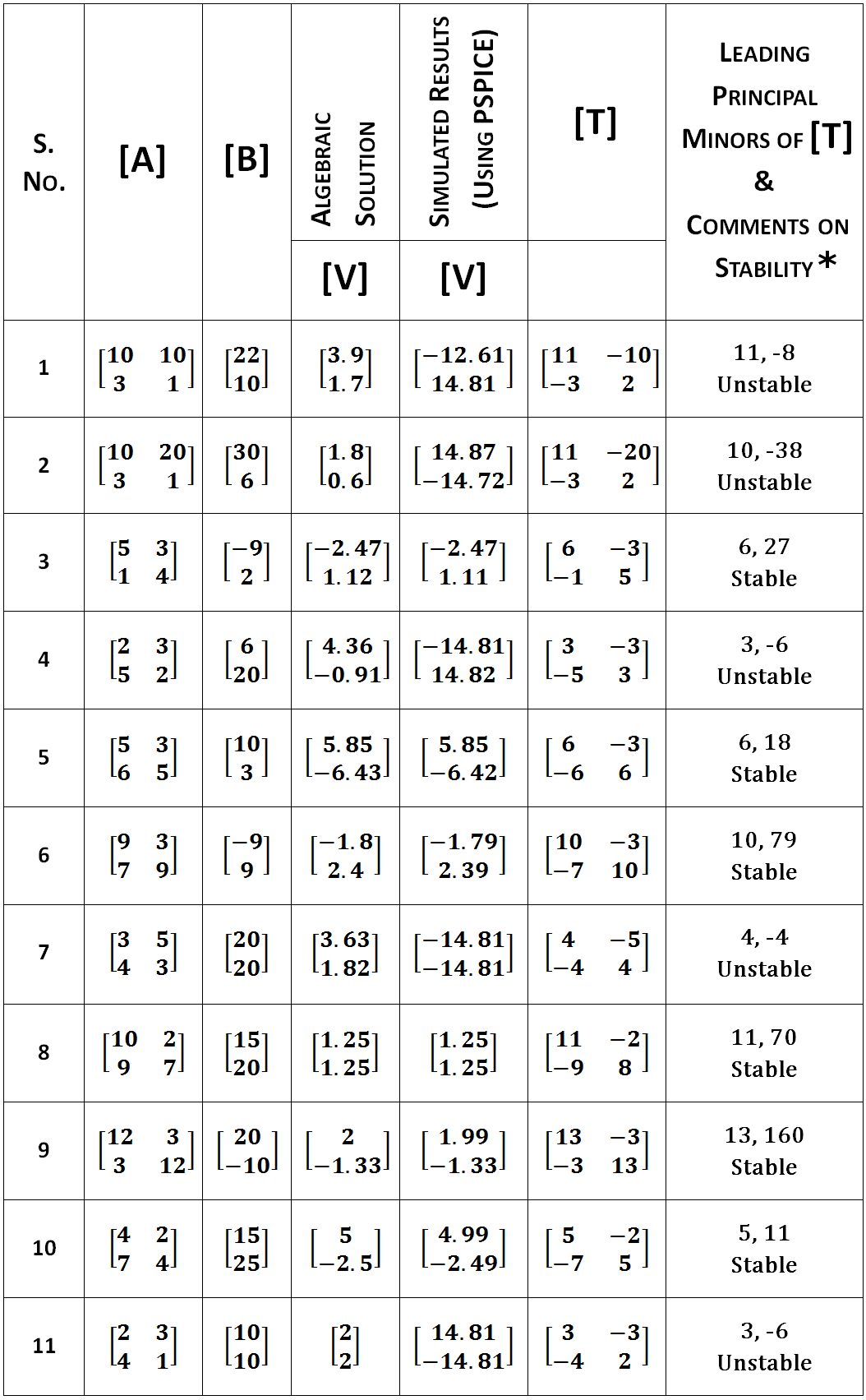}
\\ $*$the stability criteria is as elucidated in Section~\ref{AHNN}
\end{tabular}
\end{table}
\begin{table}[!htb]
\caption{Results of the AHNN used to converge to solutions of various sets of linear equations in 3 variables}
\label{tab:sle-single-layer-sim-ver-results-2}
\begin{tabular}{c}
\includegraphics[width=0.48\textwidth]{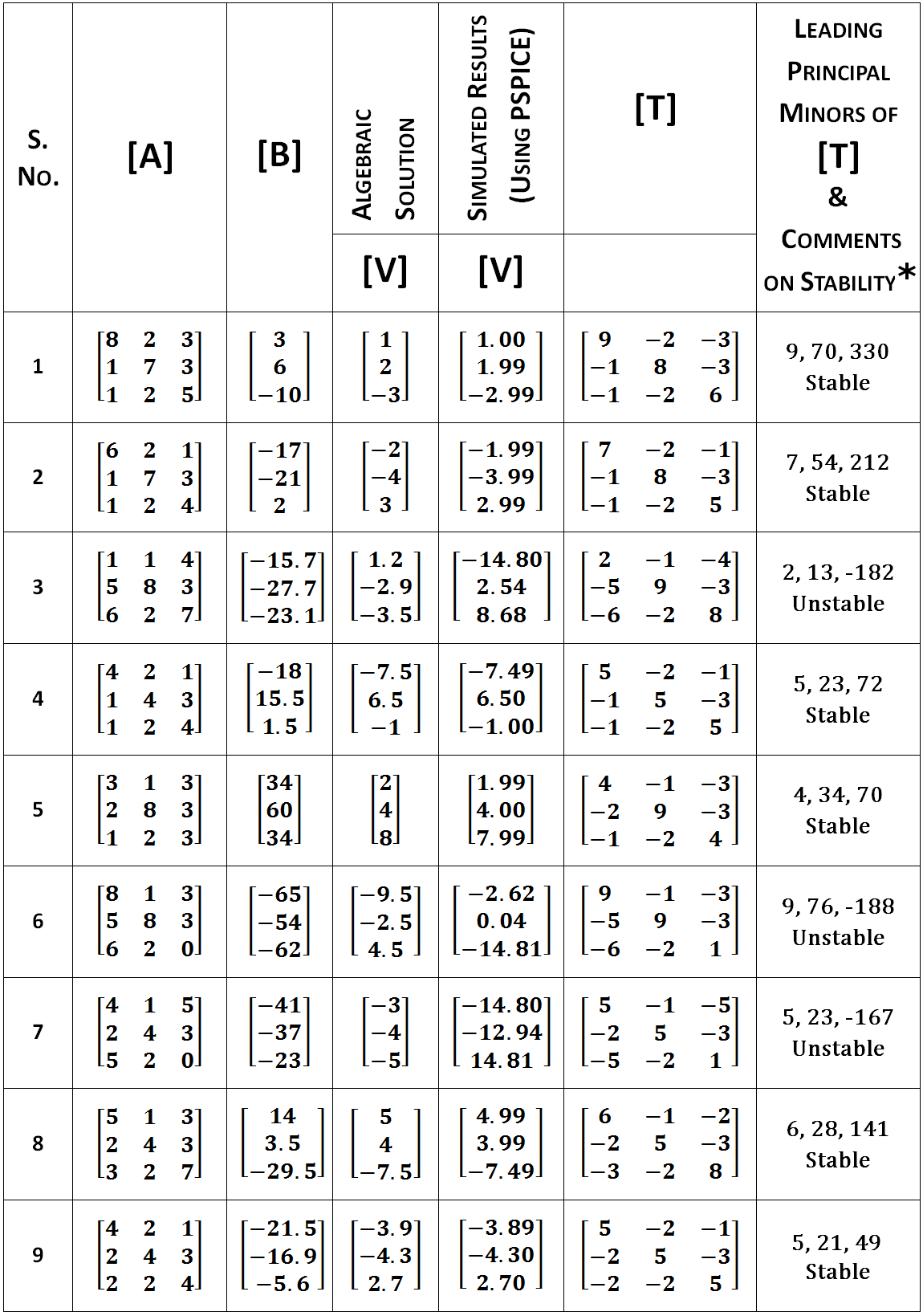}
\\ $*$the stability criteria is as elucidated in Section~\ref{AHNN}
\end{tabular}
\end{table}
\begin{table*}
\begin{center}
\caption{Results of the AHNN used to converge to solutions of various sets of linear equations in 5 variables}
\label{tab:sle-single-layer-sim-ver-results-3}
\begin{tabular}{c}
\includegraphics[width=0.95\textwidth]{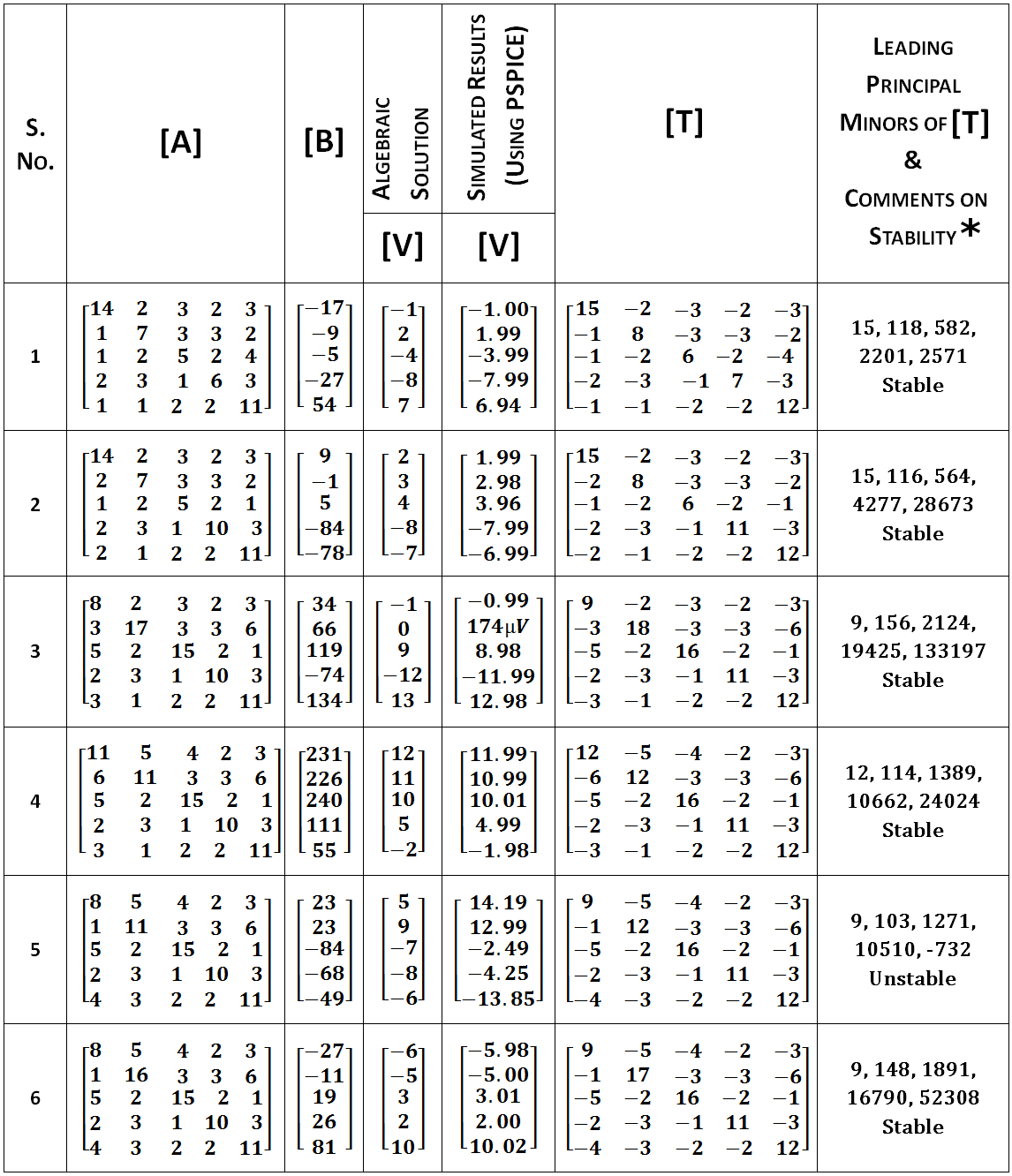}
\\ $*$the stability criteria is as elucidated in Section~\ref{AHNN}
\end{tabular}
\end{center}
\end{table*}
\begin{table*}
\begin{center}
\caption{Results of the AHNN used to converge to solutions of various sets of linear equations in 10 variables, and 20 variables. The requirement of 'diagonal dominance' is more and more evident as the size of the system of equations progresses.}
\label{tab:sle-single-layer-large-results}
\begin{tabular}{c}
\includegraphics[width=0.99\textwidth]{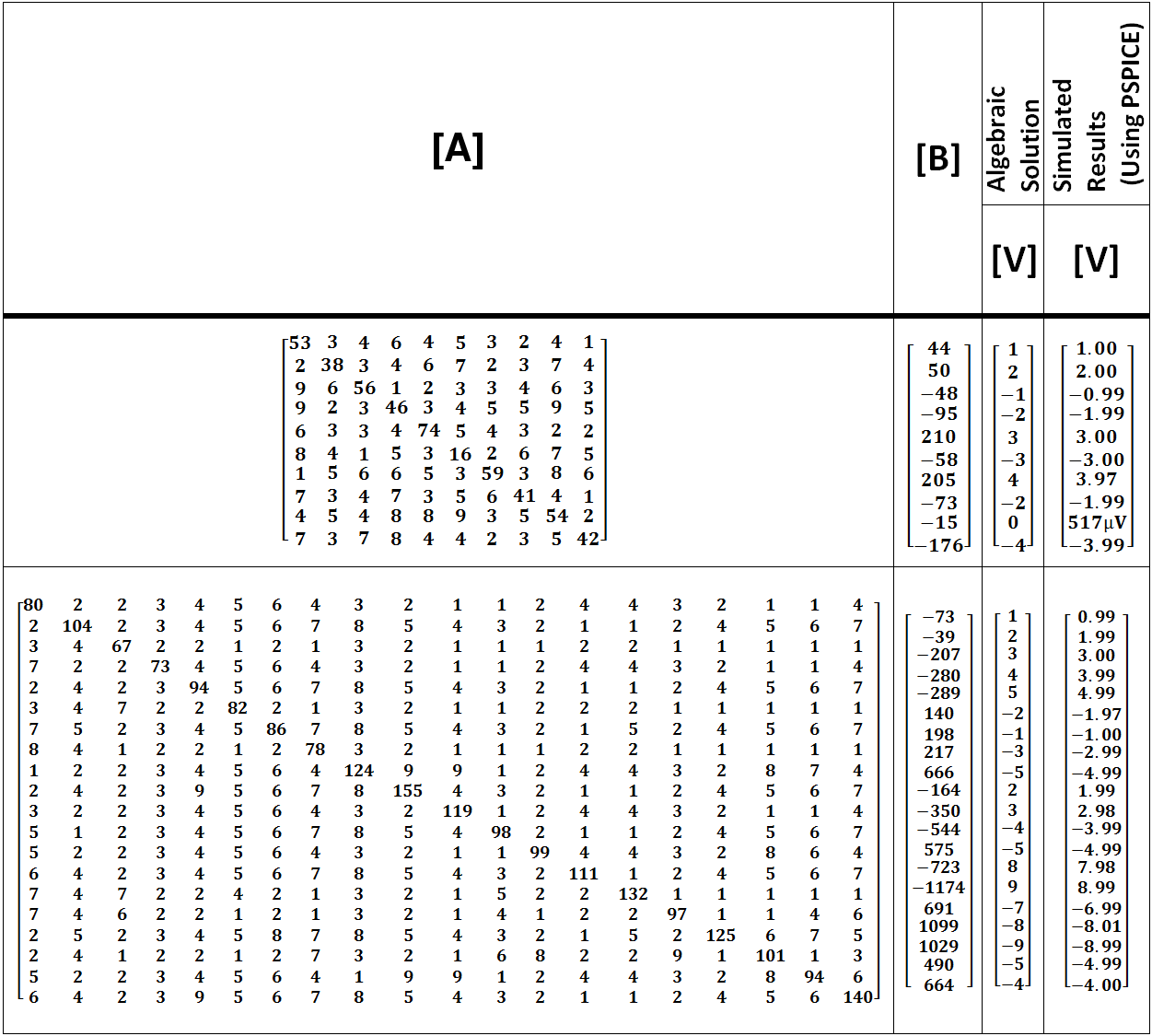}
\end{tabular}
\end{center}
\end{table*}

The proposed AHNN circuit was employed to solve (\ref{eqn:sle-single-layer-3-var}) using the resistor values provided in (\ref{eqn:sle-single-layer-3-var-rij}) and (\ref{eqn:sle-single-layer-3-var-ri}). Results of PSPICE simulations are depicted in Figure~\ref{fig:sle-single-layer-3-var-result} from where it is evident that the output node voltages (in Volts) obtained are V(1,2,3) = (1.00,$-$1.99,1.99) which are in close agreement with the exact algebraic solution of (\ref{eqn:sle-single-layer-3-var}) which is $V_1$ = 1, $V_2$ = $-$2 and $V_3$ = 2.

Next, different sets of 3 SLE in 3 unknowns were solved with the proposed AHNN. The outcomes for each of the different set of SLE, using circuit simulations, are shown in Table~\ref{tab:sle-single-layer-sim-ver-results-2}. It is seen that the circuit is indeed capable of providing correct solutions for all the cases (for which the stability criterion elaborated in Section~\ref{AHNN} is satisfied).

To test the AHNN further, different sets of 5 SLE in 5 variables were then solved with the proposed network. The results of PSPICE simulations for each of the chose set of equations are tabulated in Table~\ref{tab:sle-single-layer-sim-ver-results-3} from where it is clearly evident that the proposed AHNN  converges to correct solutions for all the chosen  sets of equations (provided that the stability criteria mentioned earlier is fulfilled).

Next, to test the scalability of the proposed AHNN for higher number of variables, the proposed circuit was then tested for the solution of SLE in 10 and 20 variables. The results as obtained from the circuit, using simulations with PSPICE, are provided in Table~\ref{tab:sle-single-layer-large-results} from where it can be seen that the AHNN does indeed outputs accurate solutions for both the large sets of SLE (10 variables and 20 variables). The voltage output plots (as obtained from PSPICE) for the AHNN configured for the 10 and 20 variable SLE are depicted in Figure~\ref{fig:sle-single-layer-10-var-result} and Figure~\ref{fig:sle-single-layer-20-var-result} respectively. From these plots, it is readily evident  that all the output voltages initially start from zero, and eventually converge to their respective `solution' values.

\section{Hardware Test Results}

\begin{figure}
\includegraphics[width=0.5\textwidth]{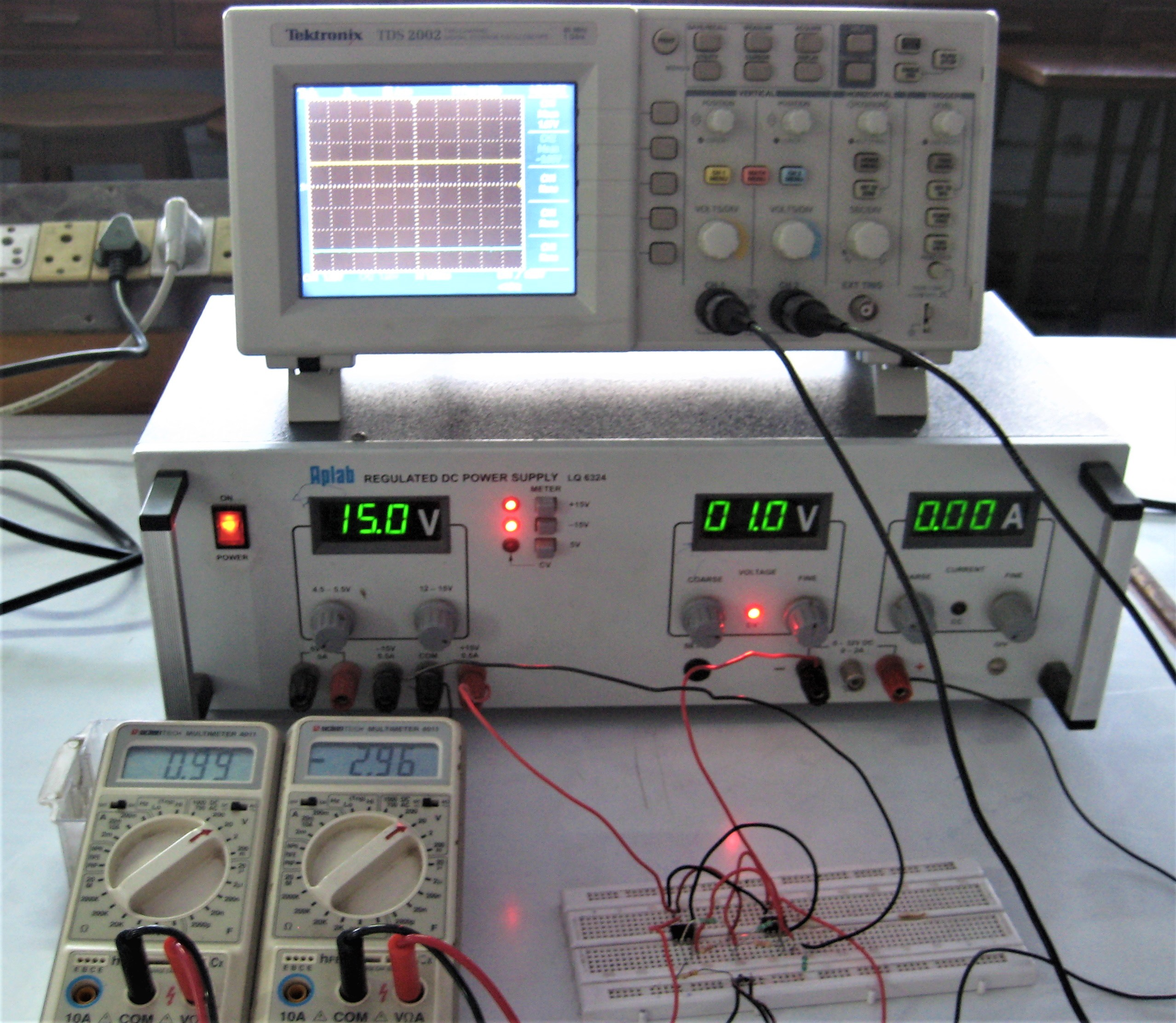}
\caption{Snapshot of the experimental affirmation of the correct working of the AHNN for a sample 2--variable problem (\ref{eqn:sle-single-layer-2-var-2})}
\label{fig:sle-single-layer-exp-result-2-var-1}
\end{figure}

Although the results of the extensive circuit simulations presented in the previous section validate the proposed AHNN, further confirmation of the network functionality is demonstrated by performing hardware realizations of the AHNN for small-sized problems, using commercially available electronic components. In addition to the  verification of expected functionality of the proposed AHNN, such actual breadboard implementations using discrete electronic  components also provide an opportunity to test the convergence of the AHNN to the desired (mathematical) solution, commencing the convergence  from diverse initial conditions (i.e. different/random internal node voltages in the circuit -- as would be the case  when the circuit is operated in a real-world scenario). The noise present in any electronic circuit (e.g. Johnson noise, Shot noise) causes the circuit to `start-up' from a random initial condition every time it is powered on.   Stock electronic components \textit{viz.} the {$\mu$}A741 operational amplifier  and axial-lead resistors have been used for the breadboard implementations.

\begin{table}
\caption{Results of experimental verification for the AHNN}
\label{tab:sle-single-layer-exp-results}
\begin{tabular}{c}
\includegraphics[width=0.48\textwidth]{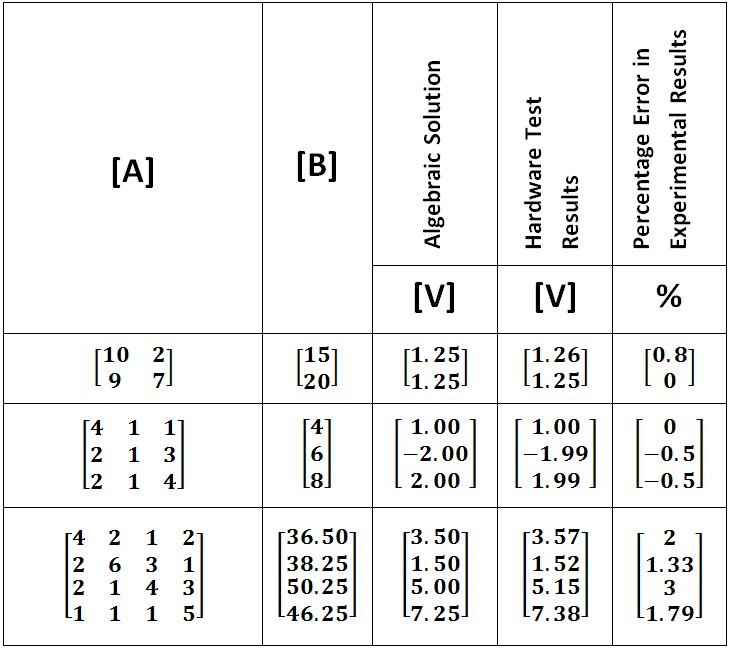}
\end{tabular}
\end{table}

The proposed AHNN was first tested by setting up the hardware for the solution of the two equations (\ref{eqn:sle-single-layer-2-var-2}). A photograph of the experimental set-up, and the obtained  results, both on a digital Cathode Ray Oscilloscope (CRO) and digital multimeters, is included in Figure~\ref{fig:sle-single-layer-exp-result-2-var-1}, from where it is readily seen that the obtained (stable, converged) values (in Volts) of the  outputs of the 2 neurons are (0.99, $-$2.96). These values when compared with the actual solution (1, $-$3) highlight the correct functionality of the breadboard realization.

\begin{equation} \label{eqn:sle-single-layer-2-var-2}
\left[ 
\begin{array}{cc}
3  &  4 \\ 
2  &  3  
\end{array}
\right]
\left[ 
\begin{array}{c}
V_1 \\ 
V_2 
\end{array}
\right]
=
\left[ 
\begin{array}{c}
-9 \\ 
-7 
\end{array}
\right]
\end{equation}

Different examples of linear equation sets (with variables ranging from 2 to 4) were then chosen and the AHNN hardware for each was then set up in the laboratory using commercially available discrete components. Each of the circuits was run 10 times so as to see whether the circuit converges to the same solution every time after being started up with varying (distinct) initial conditions (due to the inherent noise which is always present in the system and cannot be eliminated). Outcomes of the hardware implementation tests are tabulated  in Table~\ref{tab:sle-single-layer-exp-results}. It is clear from Table~\ref{tab:sle-single-layer-exp-results} that the network achieves convergence to the correct solution in all the experiments and for all the 10 trials for each experiment. The hardware solution values shown in Table~\ref{tab:sle-single-layer-exp-results} are the mean of the outputs for all 10 runs. Percentage errors for all the obtained solution values are also mentioned in Table~\ref{tab:sle-single-layer-exp-results}, and it can be seen that the hardware implementations also provide solutions very near to the mathematically exact solutions (as was the case with the PSPICE simulation results). The maximum error in the output value   is 2\%.

\section{Discussion}
\label{sle-hnn:Discussion}

This section contains a concise discussion on the applicability, hardware complexity, the strict restrictions on the values of resistors, and the scalability of the circuit.

\subsection{Applicability}

As has been discussed in previous sections, the proposed network is applicable only for those systems of equations which strictly satisfy the criterion of stability laid out in this work. Although the stability definition in this paper is presented in the form of \textbf{T} and \textbf{M} matrices, some other versions of the stability conditions are easier to comprehend.  For instance, the presence of diagonally row (or column) dominance in the interconnection matrix \textbf{W} is the simplest form amongst the several stability criteria listed in Section~\ref{AHNN}. As is well known, a given square matrix \textbf{A} can be referred to as a diagonally dominant matrix if, for each  row of \textbf{A}, the magnitude (absolute value) of the diagonal entry in that particular row is greater than, or equal to, the total of the absolute values (magnitudes) of all the other entries (i.e. the non-diagonal elements)  in that row. Algebraically this can be written in the following form. A given matrix \textbf{A} is diagonally dominant if
\begin{equation}
\label{eqn:diagonal_dominance}
    |a_{ii}| \geq \sum_{j \ne i} |a_{ij}|
\end{equation}
where $a_{ij}$ denotes the elements at the location specified by the  $i$-th row and $j$-th column.

The above phenomenon of diagonal dominance in the coefficient matrices can be readily verified by observation of Table~\ref{tab:sle-single-layer-sim-ver-results-3} and Table~\ref{tab:sle-single-layer-large-results}. It is to be noted that as the size of the coefficient matrix increases, the diagonal elements need to be more and more large in magnitude to satisfy (\ref{eqn:diagonal_dominance}). 

It is well known that several  matrices that appear in finite element methods  (FEMs) analyses are intrinsically having the diagonally dominance property. Further, it has also been established that sets of linear partial differential  equations generated from the Boundary Element Method (BEM) can be solved iteratively, with assured convergence to the accurate solution, if the BEM equations can be first recast into an analogous, diagonally dominant set of equations \cite{Urekew1992}. In the light of these applications, the proposed network assumes significance where a fast and efficient solution technique is required.

In view of the limited applicability scenarios for the circuit presented in this work, alternative analog hardware techniques (which may or may not be neural networks based) are  required for designing a generalized network which is capable of generating valid \& correct solutions for all sets of linear equations (irrespective of the satisfaction of the stability criteria as mentioned in this paper), provided that a unique (single) solution to the systems of equations under consideration exists. Some such neural circuits which are capable of this functionality, albeit at the cost of increased hardware complexity, are discussed in \cite{ansari2011dvcc,rahman2011neural,ansari2013non}.

\subsection{Hardware Complexity}

AS can be deduced from Figure~\ref{fig:sle-single-layer-full-network}, the following set of hardware is required for the proposed circuit. For solving an $n$ variables set of equations, $n$ operational amplifiers, $n^2+n$ resistors, and $n+2$ DC voltage sources are required. Amongst the $n+2$ DC voltages mentioned, $n$ sources are required for implementing the constant inputs $b_i/s_i$ for connecting to the non-inverting input terminals of the opamps, and 2 voltage sources are required for biasing the opamp ($\pm{V_m}$). In an actual integrated circuit implementation, the $n$ voltages may be generated using a potential divider type arrangement, or a multi-output reference voltage supply may be employed.

since the stability and convergence of the circuit is dependent upon the values of the resistances, it is important to have precise valued resistors (with negligible thermal drift) in a real-world implementation of the proposed circuit. This can be achieved by (i) laser trimming of the integrated resistors \cite{su2020precise}, or (ii) by using voltage-controlled electronic resistors \cite{yucel2017new}. While the former will lead to a circuit hard-wired to solve a certain set of equations, the latter may be a better choice for imparting operational flexibility to the proposed circuit.

\subsection{Scalability}

While the proposed network is theoretically able to solve $n$ linear equations in $n$ variables, provided the condition of stability is satisfied, there are going to be limits on $n$ in an actual implementation of the circuit. First, the requirement of the $n^2+n$ precise resistors puts a restriction on the extent of integration possible for a chip implementation. Secondly, the requirement of $n$ voltage supplies to provide the $b_i/s_i$ inputs to the various neurons also marks a limit that would be defined by the hardware availability. Lastly, the noise floor will also play a role in dictating the maximum number of outputs (and therefore the number of variables $n$) allowed.

\subsection{Speed}

Most of the discussion contained in the Section~\ref{sle-hnn:Circuit Simulation Results} was centered around the accuracy of the obtained solutions and not on the time taken by the circuit to converge to the solution. As can be seen from the simulation results included in Figure~\ref{fig:sle-single-layer-3-var-result} through Figure~\ref{fig:sle-single-layer-20-var-result}, the network converges to the correct solution(s) in time of the order of microseconds. This convergence time is obtained when the simulations were performed using a IC-741 PSIPCE macromodel\footnote{LM741 Macromodel available from: www.ti.com/lit/zip/snom211}. This PSPICE model was used to demonstrate the workability of the proposal, and is in no way a reflection on the actual convergence speed that will be available when modern day fast opamps (such as Linear Technology's LTC6253-7 2GHz operational amplifier\footnote{https://www.analog.com/media/en/technical-documentation/data-sheets/62537f.pdf}) are employed.

\section{Conclusion}
\label{sle-hnn:Conclusion}

An HNN-based circuit (referred to as the Asymmetric Hopfield Neural Network) was presented which was capable of providing the solution of a system of $n$ linear equations in $n$ variables. Since the weight matrix (\textbf{W}) of the AHNN is dependent on (and derived from) the coefficient matrix (\textbf{A}) of the  set of equations, it  therefore cannot be said to be axiomatically symmetric. The AHNN circuit therefore falls in the class of asymmetric HNNs \cite{Xu96, Che01}. Since the asymmetry of \textbf{W} imposes certain limitations on the stability of the AHNN, it is evident that the linear equation solver circuit proposed in this paper is amenable to a limited category of scenarios, that is only for those systems of linear equations wherein \textbf{A} satisfies the  criterion of stability as outlined in (\ref{eqn:calc-a-matrix}).
It is to be noted that since the convergence and stability of the AHNN is not universally guaranteed for all \textbf{W}, any applications of these circuits, including the equation-solver elaborated here, has a restricted applicability scenario. Alternative ANN techniques are therefore required for obtaining a generalized network which is capable of generating valid (correct) solutions for all sets of linear equations (irrespective of the presence or absence of the diagonal dominance property in \textbf{A}), provided that a unique (single) solution to that system of equations exists.


\balance
\bibliographystyle{unsrt}
\bibliography{references}

\end{document}